\documentclass[sigconf]{acmart}
\acmConference{KDDCup'25}{August 3-7 2025}{Toronto, Canada}
\AtBeginDocument{%
  }

\usepackage{pifont}
\usepackage{tcolorbox}
\usepackage{mdframed}
\begin{document}

\makeatletter
\renewcommand\@acmISBN{}
\renewcommand\@formatdoi[1]{}
\makeatother

\title{A Curriculum Learning Approach to Reinforcement Learning: Leveraging RAG for Multimodal Question Answering}


\author{Chenliang Zhang}
\email{zhangchenliang02@meituan.com}
\authornote{Both authors contributed equally to this research.}
\affiliation{%
  \institution{Meituan}
  \city{Shanghai}
  \country{China}
}

\author{Lin Wang}
\authornotemark[1]
\email{wanglin84@meituan.com}
\affiliation{%
  \institution{Meituan}
  \city{Shanghai}
  \country{China}
}

\author{Yuanyuan Lu}
\email{luyuanyuan04@meituan.com}
\authornotemark[1]
\authornote{Corresponding Author.}
\affiliation{%
  \institution{Meituan}
  \city{Shanghai}
  \country{China}
}

\author{Yusheng Qi}
\email{qiyusheng@meituan.com}
\authornotemark[1]
\affiliation{%
  \institution{Meituan}
  \city{Shanghai}
  \country{China}
}

\author{Kexin Wang}
\email{wangkexin23@meituan.com}
\authornotemark[1]
\affiliation{%
  \institution{Meituan}
  \city{Shanghai}
  \country{China}
}

\author{Peixu Hou}
\email{peixu.hou@meituan.com}
\affiliation{%
  \institution{Meituan}
  \city{Shanghai}
  \country{China}
}

\author{Wenshi Chen}
\email{wenshi.chen@meituan.com}
\affiliation{%
  \institution{Meituan}
  \city{Shanghai}
  \country{China}
}

\begin{abstract}
This paper describes the solutions of the Dianping-Trust-Safety team for the META CRAG-MM challenge. The challenge requires building a comprehensive retrieval-augmented generation system capable for multi-modal multi-turn question answering. The competition consists of three tasks: (1) answering questions using structured data retrieved from an image-based mock knowledge graph, (2) synthesizing information from both knowledge graphs and web search results, and (3) handling multi-turn conversations that require context understanding and information aggregation from multiple sources. For Task 1, our solution is based on the vision large language model, enhanced by supervised fine-tuning  with knowledge distilled from GPT-4.1. We further applied curriculum learning strategies to guide reinforcement learning, resulting in improved answer accuracy and reduced hallucination. For Task 2 and Task 3, we additionally leveraged web search APIs to incorporate external knowledge, enabling the system to better handle complex queries and multi-turn conversations. Our approach achieved 1st place in Task 1 with a significant lead of 52.38\%, and 3rd place in Task 3, demonstrating the effectiveness of the integration of curriculum learning with reinforcement learning in our training pipeline.

\end{abstract}




\maketitle

\section{Introduction}
Recent advances in Vision Large Language Models (VLLMs) have significantly enhanced the capabilities of multi-modal systems, enabling more sophisticated visual question answering (VQA) and multi-modal understanding. However, despite these improvements, VLLMs remain susceptible to generating hallucinated responses, particularly when faced with queries involving rare entities or requiring complex reasoning that spans recognition, OCR, external knowledge integration, and answer generation. To address these limitations, the Retrieval-Augmented Generation (RAG) paradigm has emerged, allowing models to ground their responses in external knowledge sources by retrieving and synthesizing relevant information from structured knowledge graphs and unstructured web content.

The MM-RAG QA competition presents a rigorous benchmark for evaluating the next generation of multi-modal RAG systems. The challenge is structured around three progressively complex tasks: (1) leveraging structured data from image-based knowledge graphs to answer factual and recognition-based questions, (2) synthesizing information from both knowledge graphs and web search results to answer more complex, knowledge-intensive queries, and (3) engaging in multi-turn conversations that require context tracking and information aggregation across multiple sources and dialogue turns. The benchmark further categorizes questions into simple recognition, simple knowledge, multi-hop, comparison, aggregation, and reasoning types, reflecting real-world demands on robust multi-modal QA systems.

In this competition, our team developed a comprehensive MM-RAG solution built on a VLLM foundation. Our approach integrates supervised fine-tuning via knowledge distillation from GPT-4.1 and introduces curriculum learning to guide reinforcement learning, thereby enabling the model to progressively master increasingly complex tasks. For tasks requiring external knowledge, we additionally incorporate web search APIs to enhance retrieval and synthesis capabilities. Our system demonstrated strong performance, achieving 1st place in Task 1 with a significant 52.38\% lead and 3rd place in Task 3, underscoring the effectiveness of combining curriculum learning with reinforcement learning for advancing multi-modal question answering. Our code is available on Gitlab \footnote{https://gitlab.aicrowd.com/ysQi/meta-comprehensive-rag-benchmark-starter-kit} .

The remainder of this paper is organized as follows. In Section 2, we review related work on visual question answering. Section 3 presents our methodology, including supervised fine-tuning, reinforcement learning with curriculum learning, and retrieval-augmented generation module. Section 4 describes our experimental setup and analyzes the performance of our approach. Finally, Section 5 concludes the paper and discusses potential future work.

\section{Related Work}

\subsection{Visual Question Answering}

Visual Question Answering as a cross-modal task combining computer vision and natural language processing has witnessed significant advancement in recent years. Antol et al.~\cite{antol2015vqa} first introduced a large-scale VQA dataset, laying the foundation for research in this field. The architecture of VQA systems evolved from using simple fusion operations to more complex methods including CNN~\cite{ma2016} and LSTM~\cite{malinowski2015}. Visual feature extraction transitioned from grid-based approaches using CNN architectures like ResNet~\cite{he2016} to object-based methods~\cite{anderson2018bottom}, and finally to ViT-based patch encoders~\cite{dosovitskiy2020}. For language encoding, the evolution progressed from simpler representations to Transformer-based architectures represented by BERT~\cite{devlin2019bert}.

The advent of VLLMs has revolutionized the VQA field. Early notable works include ViLBERT~\cite{lu2019vilbert} and VL-BERT~\cite{su2019}, which extended BERT architecture to handle multi-modal inputs. LXMERT~\cite{tan2019} proposed cross-modality pre-training to learn joint visual and language representations. More recent approaches shifted toward unified Transformer frameworks and contrastive learning methods exemplified by CLIP~\cite{radford2021learning}, which demonstrated powerful zero-shot learning capabilities. Contemporary VLLMs employ various pre-training strategies including masked modeling techniques and generative approaches. The latest trend involves fine-tuning existing large language models (LLMs) like Llama~\cite{meta2023}, Qwen~\cite{bai2023}, and Falcon using adapters that map visual tokens to language tokens, significantly reducing training resources while maintaining state-of-the-art performance across various benchmarks~\cite{liu2023a, zhu2023}.

\section{Methodology}
\begin{figure*}[!htbp]
    \centering
    \includegraphics[width=0.8\linewidth]{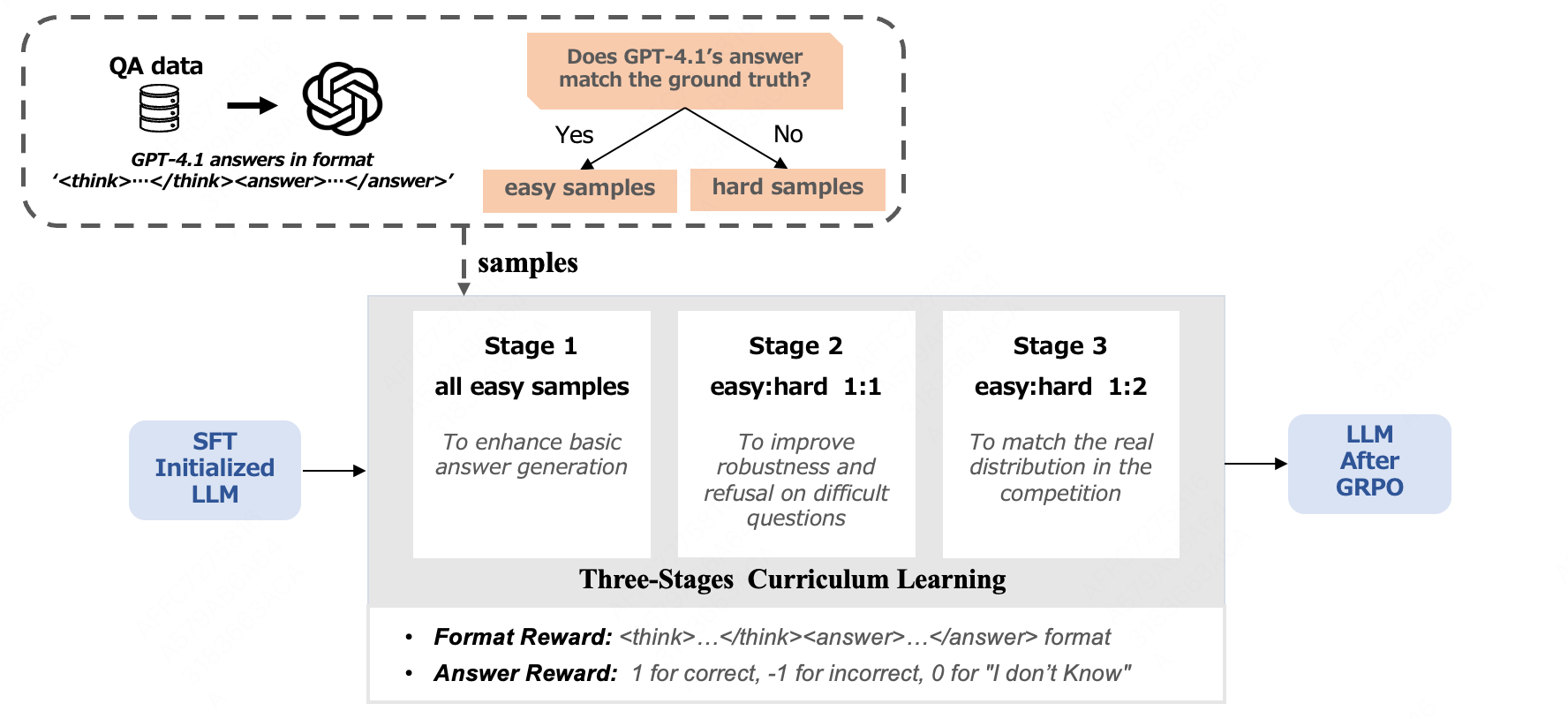}
    \caption{An illustration of curriculum learning.}
    \label{fig:curriculum}
\end{figure*}

Our approach is based on three key components: supervised fine-tuning via knowledge distillation, reinforcement learning with curriculum learning, and a retrieval-augmented generation module for multi-source augmentation.

\subsection{Supervised Fine-tuning}

Given the evaluation protocol imposes a severe penalty for hallucinations (with a score of -1 in evaluation), our SFT strategy is designed not only to improve answer accuracy but also to lay the foundation for subsequent control of hallucinations. Since our base model, Llama 3.2–Instruct–11B, is a relatively compact vision-language model, its initial capabilities in image understanding, reasoning, and confidence estimation are limited.

During supervised fine-tuning, we focus on enhancing the model’s ability to generate accurate chain-of-thought (CoT) style answers in the required \texttt{<think></think>} and \texttt{<answer></answer>} format, rather than directly controlling hallucinations at this stage. To construct high-quality SFT training data, we distill both answers and detailed reasoning processes from stronger models such as GPT-4.1~\cite{achiam2023gpt}. In our prompting strategy, GPT-4.1 is provided with the ground-truth answer and is instructed to decide whether the question is answerable. If so, it generates both the answer and a corresponding CoT; if not, it returns an explicit refusal and the reason for being unable to answer. This ensures that the SFT data reflects both high-quality reasoning and the ability to abstain when appropriate. We use the following prompt to generate responses with GPT-4.1:

\begin{mdframed}[linewidth=1pt,skipabove=10pt,skipbelow=10pt]
\begin{footnotesize}
\begin{verbatim}
[{"role": "user", "content": "You are an intelligent assistant 
that can answer users' questions based on the pictures they provide.
You need to put your thinking/reasoning  process into <think>xxx
</think> and put the answer or conclusion into <answer>xxx</answer>. 
Please avoid wrong answers and hallucinations. When you can get 
a high-confidence answer, <answer>xxx </answer> is your answer; 
When your current knowledge is not enough for you to answer the 
question, <answer>xxx</answer> should be  <answer>I don't know, 
please add xxx information</answer>; don't give ambigious answer. 
Your response should strictly follow the format of <think>xxx
</think>, <answer>xxx</answer>. "}
{"role": "user", "content":"Please answer the following 
question based on the provided image: Please answer the following 
question based on the provided image: {query_str}"}, 
{"role": "user", "content": "The answer is: {answer_str}"}, 
{"role": "user", "content": [{"type": "image"}]}]
\end{verbatim}
\end{footnotesize}
\end{mdframed}
It is important to note that during the SFT of the Llama model, the answer is not provided.

In addition, for the integrated auxiliary search tools such as image search and web search, the model needs to acquire the ability to invoke these tools. Specifically, we leverage more powerful models (such as GPT-4.1) to generate explicit CoT outputs for tool usage, and use these data to fine-tune the Llama Vision model. This enables the model to, given an image and a question, recognize what knowledge it lacks and determine which search tool to use and what search query to issue. Concretely, if the model needs to use image search, it should output the coordinates of the relevant sub-region of the original image; if web search is required, it should output one or more text queries. Our experiments show that the use of image search is not effective, and this will be discussed in detail in subsequent sections. For the additional knowledge retrieved via search, we concatenate it with the original question to help the model generate the final answer.  We use the following prompt to generate querys: 
\begin{mdframed}[linewidth=1pt,skipabove=10pt,skipbelow=10pt]
\begin{footnotesize}
\begin{verbatim}
[{"role": "user", "content": "You are a RAG search assistant. 
For the given question and picture, understand and analyze the 
question, abstract the specific question, and analyze the purpose 
of the question.Please summarize a query from the question 
( the query will be used to search for relevant information 
in the text rag later) to help the subsequent small model answer 
the current question. Your query should be detailed and concise, 
and cover the key information to solve the current problem, such
as: <BMW i5 2025 price><Audi q5l engine><Dragon fruit China distri-
bution><Azalea flowering period><White House construction date>
<Eiffel Tower designer> etc. Please write the reasoning process 
in <think></think> and the query in <search></search>.
-- Your reasoning process should be detailed and step-by-step. 
-- When the current problem is simple recognition and simple calcu-
lation, no additional search information is required, and your 
query can be empty. That is, <think>reasoning</think><search>
</search>, it must be strictly  in this form. In most cases, you 
should use query first, because this information can help the 
downstream small model answer questions and effectively avoid 
the illusion of the small model."},
{"role": "user", "content":"The question is: {query_str}"}, 
{"role": "user", "content": [{"type": "image"}]}]
\end{verbatim}
\end{footnotesize}
\end{mdframed}
\vspace{5pt}

This is the prompt to generate responses using retrieval informaton:

\begin{mdframed}[linewidth=1pt,skipabove=10pt,skipbelow=10pt]
\begin{footnotesize}
\begin{verbatim}
[{"role": "user", "content": "You are an intelligent assistant 
and can answer users' questions based on the pictures provided 
by users. Please avoid giving wrong answers and having halluci-
nations.When you can obtain a highly reliable answer, return the 
answer directly; When your existing knowledge is insufficient to 
answer the question, your answer should be 'I don't know.'' For 
Yes/No questions, the answers should start with 'Yes' or 'NO', and 
no ambiguous answers should be given. In addition, the web search 
content will also be provided for your reference. Please carefully 
verify whether the provided search information is reliable enough 
to help you answer correctly. The search information is 
{search_result}."}
{"role": "user", "content":"The question is: {query_str}"}, 
{"role": "user", "content": [{"type": "image"}]}]
\end{verbatim}
\end{footnotesize}
\end{mdframed}
\vspace{5pt}

After SFT, we observe a notable improvement in the model’s accuracy and its preliminary ability to distinguish answerable from unanswerable questions. To maximize training efficiency and training throughput, we adopt the LoRA~\cite{hu2022lora} technique for fine-tuning. The explicit control of hallucinations is deferred to the subsequent reinforcement learning stage, where reward design and curriculum learning play a central role.

\subsection{Reinforcement Learning with Curriculum Learning}

In the reinforcement learning (RL) stage, our main objective is to optimize the reasoning capabilities of the model: not only its ability to produce correct answers, but also its capacity to reason about whether a question can be reliably answered, thus minimizing hallucinations. To this end, we carefully designed both the prompting strategy and the reward function.

We adopt the VisualRFT framework~\cite{liu2025visual}, which supports efficient rollout generation, relative advantage computation, reference model loss calculation, and policy updates, and we extend it to fully support the Llama architecture. The underlying algorithm is based on Grouped Relative Policy Optimization (GRPO), with a composite reward function comprising both format and answer rewards. Specifically, the format reward encourages outputs in the required \texttt{<think></think>} and \texttt{<answer></answer>} structure, while the answer reward assigns 1 for correct answers, 
0 for missing, and -1 for incorrect or hallucinated responses, as judged by a GPT-4o mini evaluator.

A key challenge in this setting is that, under the strict evaluation protocol, the model can easily fall into a “reward black hole”—learning to refuse answering almost all questions to avoid penalties, resulting in a missing rate as high as 90\%. To address this, we introduce a curriculum learning strategy~\cite{bengio2009curriculum}, enabling the model to learn in a staged, progressively difficult manner.

As illustrated in Figure 1, we categorize training samples into "easy" and "hard" groups using GPT-4o mini: samples that GPT-4.1 answers correctly (under the same evaluation protocol as the leaderboard) are labeled as easy (about 1,300 samples in our training set), while those it cannot answer are labeled as hard. The RL process proceeds in three stages:
\begin{itemize}
\item \textbf{Stage 1:} Training only on easy samples to enhance the model's basic answer generation and reasoning ability.
\item \textbf{Stage 2:} Training on a 1:1 mix of easy and hard samples to improve robustness and refusal ability on difficult questions.
\item \textbf{Stage 3:} Training on a distribution that matches the real competition (approximately 1:2 easy to hard), allowing the model to adapt to the real-world scenario.
\end{itemize}

The introduction of curriculum learning greatly stabilizes the RL process. In the early stages, the model reliably acquires strong reasoning and answer generation skills without prematurely converging to high refusal rates. In later stages, it learns to appropriately refuse unanswerable questions, effectively reducing hallucinations while maintaining a high answer rate. Overall, this staged approach enables the model to achieve a balanced trade-off between accuracy and hallucination, resulting in superior performance in the competition setting.

In addition, the model receives different inputs during inference across the three tasks. In Task 1, only the original image and question are provided. In Task 2, additional auxiliary information is included. Task 3 further incorporates historical dialogue information. To accommodate these task variations, we employ different adaptation strategies during reinforcement learning training. Specifically, for Task 1, the model is trained using only the original image and question as input. For Task 2, auxiliary information is added to the input. For Task 3, both auxiliary information and randomly selected historical dialogue are concatenated into the input. These tailored training strategies enable the model to adapt effectively to the requirements of the three distinct tasks.

During the GRPO phase, while focusing on enhancing answer accuracy, there is an observed trade-off where the model's ability to generate queries might slightly diminish. This is acceptable because our primary goal is to improve the accuracy of answers and effectively control hallucinations. Despite this trade-off, the model retains enough query generation capability to support complex question answering.

\subsection{Retrieval-Augmented Generation }

\begin{figure*}[htbp]
    \centering
    \includegraphics[width=0.8\linewidth]{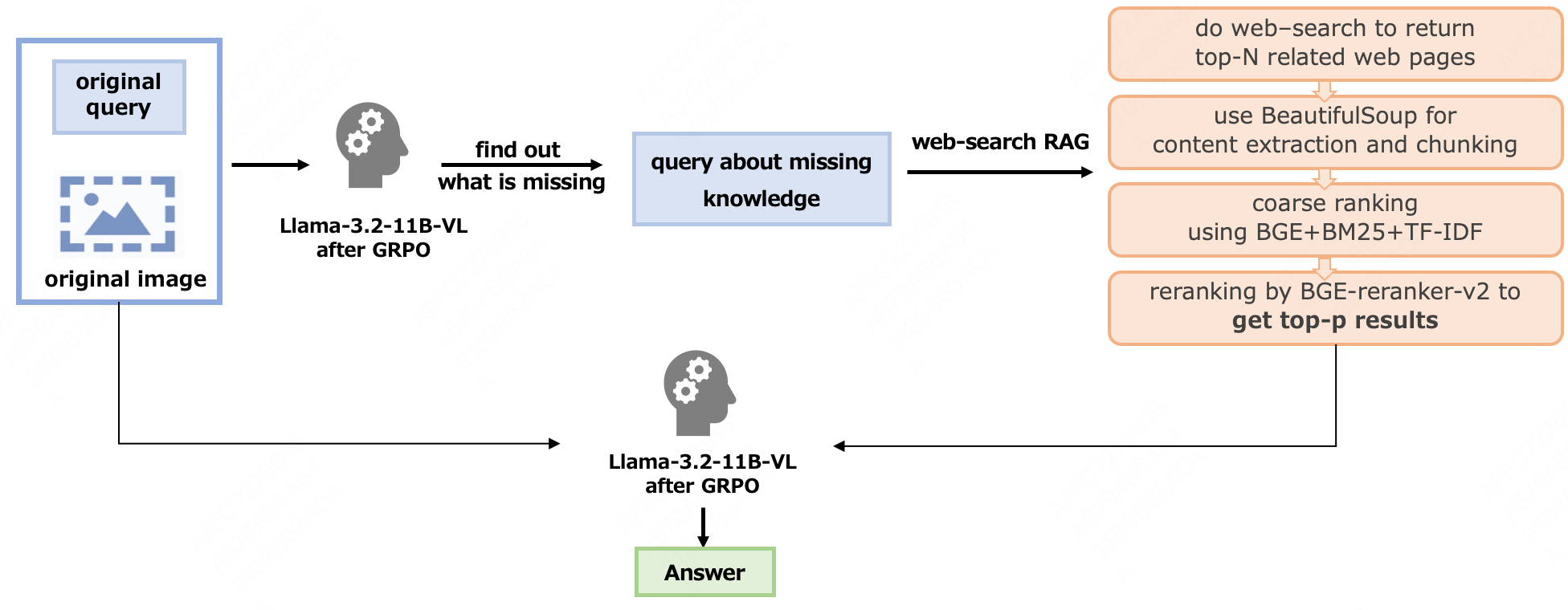}
    \caption{An illustration of RAG system.}
    \label{fig:rag}
\end{figure*}

RAG is a technique that enhances large language models by retrieving relevant knowledge to supplement their input. In this challenge, Task 1 allows the use of image search, while Task 2 and Task 3 additionally permit web search. Traditional RAG methods typically encode the original query and retrieve the top-K most relevant knowledge snippets to concatenate with the model input. However, this approach often loses important cross-modal semantic information. For example, given the query "Who is the CEO of the company that manufactures this?" and an image of a BMW i3, the ideal search query should be "Who is the CEO of the company that manufactures the BMW i3?" or simply "Who is the CEO of BMW?" rather than directly using the original query.

Inspired by Deepresearcher~\cite{zheng2025deepresearcher}, we decided to allow the model to determine what additional knowledge is needed for the current question. We informed the model that two tools are available: image search and web search. For image search, the query should be a sub-image; for web search, the query should be textual. However, during our experiments, we observed two main issues: (1) the Llama Vision model was not well-suited for object detection tasks; (2) image search introduced significant noise, even after manual sub-image extraction. These issues resulted in highly unstable model performance when using image search, sometimes even worse than not using it at all. As a result, we ultimately decided to abandon image search and rely solely on web search.

To better leverage web search, we developed a dedicated retriever. Specifically, for a given query, the retriever uses web search to obtain k web pages, extracts the main content using BeautifulSoup, and segments both the main text and summary. We perform multi-stage retrieval: initial coarse ranking is done using a combination of bge-large-en-v1.5 embeddings, BM25, and TF-IDF, followed by re-ranking with bge-reranker-v2-m3 to select the top-k results. The actual k value used for both web retrieval and the final answer selection is 5.
\begin{itemize}
    \item \textbf{Single-source augmentation (Task 1):} Since only image search is allowed in Task 1, we did not use any external knowledge. The model input consists solely of the original question and the image.
    \item \textbf{Multi-source augmentation (Task 2):} In Task 2, web search is permitted, so we divided the task into two stages. In the first stage, the model receives the original question and image, and determines whether additional knowledge is required to answer the question. If so, the model generates one or more textual queries; otherwise, it outputs nothing. These queries are then processed by the retriever to obtain the top-k results. In the second stage, the original question, image, and the retrieved top-k results are concatenated as input to the model for final answer generation. The complete workflow is illustrated in Figure 2.
    \item \textbf{Multi-turn augmentation (Task 3):} The knowledge augmentation process in Task 3 is similar to that of Task 2, with the main difference being that Task 3 involves multi-turn dialogue, as opposed to the single-turn setting in Task 2. Thus, we provide historical questions and answers as part of the reference information. The VLLM itself determines whether and how to use this historical data, allowing it to leverage past interactions to potentially enhance its understanding and response quality.
\end{itemize}
This flexible tool-usage strategy enables the model to acquire relevant knowledge according to its actual needs, thereby maximizing its exploration capabilities within limited resources and minimizing the introduction of noise.

\section{Experiments}

\subsection{Experiment Settings}
\subsubsection{Dataset} 
The CRAG-MM dataset \cite{wang2025crag} is designed to benchmark multi-modal, retrieval-augmented question answering systems and consists of three main components: an image set, a collection of question-answer (QA) pairs, and retrieval contents.

CRAG-MM includes two categories of images: (1) egocentric images captured using Ray-Ban Meta Smart Glasses, and (2) normal images collected from various public sources. This diversity ensures coverage of both first-person and conventional perspectives.

The dataset covers 13 domains, such as Books, Food, Math \& Science, Shopping, Animal, and Vehicles, among others. QA pairs are annotated with four question types: simple recognition and simple knowledge, multi-hop, comparison and aggregation, and reasoning. Both single-turn and multi-turn conversation data are included. For single-turn dialogues, the validation set contains 1,936 samples and the public test set contains 1,938 samples. For multi-turn dialogues, the validation set contains 586 samples and the public test set contains 587 samples.To facilitate local development and evaluation, we merged all available single-turn validation and public test samples and randomly split approximately 15\%  as our local validation set.

To facilitate retrieval-augmented generation, CRAG-MM provides two mock retrieval APIs:
\begin{itemize}
\item \textit{Image Search API:} Given an input image, this API returns similar images along with structured metadata from a mock knowledge graph. For example, querying with a landmark image retrieves visually similar images and their associated metadata.
\item \textit{Text-Based Web Search API:} Given a text query, this API returns a set of relevant web pages, including URLs, page titles, snippets, and last updated times.
\end{itemize}
Both APIs are designed to include hard negative samples, simulating real-world retrieval noise and enhancing the challenge for question answering systems.
\subsubsection{Metrics}

We employ a comprehensive set of evaluation metrics to assess the performance of MM-RAG systems, focusing on answer correctness, informativeness, and reliability. The four primary metrics are accuracy, missing rate, hallucination rate, and truthfulness score, with the latter serving as the final ranking criterion.
\begin{itemize}
\item \textit{Accuracy:}
Accuracy measures the proportion of answers that are fully correct. During the automatic evaluation phase, an answer is labeled as "Perfect" if it completely and correctly addresses the question, and accuracy is computed as the percentage of “Perfect” responses among all evaluated examples.

\item \textit{Missing Rate:}
The missing rate quantifies the proportion of questions for which the system fails to provide an answer or responds with statements such as "I don't know" or "I'm sorry I can't find...". These responses are labeled as "Missing" in both automatic and human evaluations.

\item \textit{Hallucination Rate:}
Hallucination is defined as the generation of incorrect or irrelevant information that is not supported by the provided image or retrieved knowledge. Answers labeled as "Incorrect" are considered hallucinations, and the hallucination rate is calculated as the proportion of such answers among all responses.

\item \textit{Truthfulness Score:}
The truthfulness score is the primary metric for system ranking. For single-turn QA, each answer is assigned a score: 1.0 for "Perfect", 0.0 for "Missing", and -1.0 for "Incorrect". In the human evaluation phase, an additional "Acceptable" category is introduced for answers that are useful but contain minor, non-harmful errors, and these are scored as 0.5. The truthfulness score is computed as the average score across all examples in the evaluation set for each system.
\end{itemize}
For multi-turn QA, following~\cite{xu2023crag}, a conversation is terminated if two consecutive answers are labeled as "Incorrect", and all subsequent answers are considered "Missing". The average score across all multi-turn conversations is then reported as the final truthfulness score.

\subsection{Overall Performance}

\begin{table}[!htbp]
\centering
\caption{Performance of our team on phase2 leaderboard and manual annotation}
\begin{tabular}{l ccc}
\toprule
Task           & Task 1 & Task 2 & Task 3 \\
\midrule
auto-evaluation rank  & \textbf{1}     & \textbf{2}     & \textbf{3}     \\ 
manual annotation rank    & \textbf{1}    & -     & \textbf{3}     \\ 
\midrule
Accuracy           &  0.181     &  0.247     &   0.258    \\ 
Missing            &  0.711     &  0.625     & 0.602      \\ 
Hallucination      &  0.108     & 0.128      & 0.140      \\ 
Truthfulness Score &  0.073     & 0.119      & 0.118      \\ 
\bottomrule
\end{tabular}
\end{table}

\begin{table*}[!htbp]
\centering
\caption{Ablation results on single-turn local evaluation set.}
\begin{tabular}{cccccccc}
\toprule
SFT & RL & CL & RAG & Accuracy & Missing & Hallucination & Truthfulness Score \\
\midrule
 \ding{55}  &  \ding{55} & \ding{55} &  \ding{55}  &  0.197 & 0.085 & 0.718 & -0.520 \\
  \ding{51}  & \ding{55} &  \ding{55}  &  \ding{55}   &  0.262 & 0.237 & 0.501 & -0.238 \\
  \ding{51}  & \ding{51} & \ding{55} &  \ding{55}   &   0.059 & 0.909 & 0.032   & 0.027        \\
  \ding{51}  &  \ding{51}  &  \ding{51}  &   \ding{55}  &  0.190 & 0.713 & 0.097  & 0.093     \\
 \ding{51}   &  \ding{51}  &  \ding{51}  &  \ding{51}   & 0.282 & 0.587 & 0.131     & 0.151   \\
\bottomrule
\end{tabular}
\end{table*}

Table~1 presents the overall performance of our system across the three competition tasks. The \textbf{auto-evaluation rank} corresponds to the real-time leaderboard, which is updated automatically based on system outputs using predefined evaluation scripts. In this setting, our method consistently ranked among the top teams, achieving \textbf{1st place} in Task~1, \textbf{2nd place} in Task~2, and \textbf{3rd place} in Task~3.

For the \textbf{manual annotation rank}, the top-20 teams from the leaderboard were selected for further evaluation through human annotation, which determines the final official ranking. In this stage, our system achieved \textbf{1st place} in Task~1 and \textbf{3rd place} in Task~3, demonstrating strong performance in both single-source and multi-turn QA scenarios. These results highlight the robustness and reliability of our approach.

Across all tasks, our system demonstrated competitive accuracy, low hallucination rates, and high truthfulness scores, validating the effectiveness of our curriculum learning and reinforcement learning framework for multi-modal question answering.

\subsection{Ablation Study}

We conducted an ablation study to evaluate the contribution of each component in our pipeline: Supervised Fine-Tuning (SFT), Reinforcement Learning (RL), Curriculum Learning (CL), and Retrieval-Augmented Generation (RAG). The results on the single-turn local evaluation set are presented in Table~2.

Our base model, Llama 3.2–11B-Vision-Instruct, without any fine-tuning or adaptation, demonstrates poor performance on the benchmark, with an accuracy of only 0.197 and a very high hallucination rate of 0.718. The resulting truthfulness score is as low as 
-0.520, reflecting the model's inability to distinguish answerable from unanswerable questions and the strong penalty for hallucinations.

Applying SFT, where we distill answers and reasoning traces from stronger models , leads to a substantial improvement: accuracy rises to 0.262, and the hallucination rate drops to 0.501. The model becomes more capable of generating correct answers and following the required reasoning format, although hallucination remains a challenge.

Introducing reinforcement learning without curriculum learning leads to an unstable training process. While RL alone is effective in reducing hallucinations by encouraging the model to refuse to answer uncertain questions, this comes at the significant cost of drastically reduced accuracy (dropping to 0.059) and an extremely high missing rate (0.909). Nevertheless, it is worth noting that, for the first time, the truthfulness score becomes positive (0.027), indicating that the model is better aligned with the competition’s ranking metric by avoiding penalized hallucinated answers—even though this is achieved by sacrificing most opportunities to answer.

Incorporating curriculum learning into the RL process addresses this issue by gradually increasing the difficulty of training samples. The model first learns from easy questions before being exposed to harder ones. This staged approach enables the model to maintain a much higher accuracy (0.190) and a substantially lower missing rate (0.713) compared to RL without curriculum. The truthfulness score also improves to 0.093, demonstrating a better balance between answer generation and refusal.

Finally, integrating the RAG component allows the model to utilize external knowledge, further boosting accuracy to 0.282 and improving the truthfulness score to 0.151. The hallucination rate is also reduced to 0.131, indicating that retrieval-augmented information helps the model answer more complex or knowledge-intensive questions while maintaining reliability.

Overall, the ablation results clearly demonstrate that each component—particularly curriculum learning and retrieval augmentation—plays a crucial role in achieving a robust trade-off between accuracy, missing rate, and hallucination, leading to the best overall truthfulness score in the final system configuration.

\subsection{Analysis of Curriculum Learning Stages}

\begin{table}[!htbp]
\centering
\caption{Performance of three curriculum learning stages on single-turn local evaluation set.}
\begin{tabular}{l cccc}
\toprule
   & Accuracy & Missing & Hallucination & Truthfulness Score \\
\midrule
stage 1                       & 0.349             & 0.056                                                  & 0.595                                                        & -0.246             \\
stage 2                       & 0.338             & 0.360                                                  & 0.302                                                        & 0.036              \\
stage 3                       & 0.282             & 0.587                                                  & 0.131                                                        & 0.151         \\
\bottomrule
\end{tabular}
\end{table}

To further analyze the effectiveness of curriculum learning, we evaluated the model at each stage of the curriculum (Table~3). In the first stage, training on easy samples enables the model to develop basic reasoning and QA abilities, reflected by a relatively high accuracy (0.349) but also a high hallucination rate (0.595), as the model tends to answer even when uncertain. In the second stage, introducing difficult samples improves the model's ability to distinguish answerable from unanswerable questions, leading to a substantial reduction in hallucination (0.302) and a more balanced performance. In the third stage, the model is exposed to the real distribution of question difficulties. This enables autonomous learning and further refinement, resulting in the best overall truthfulness score (0.151). This staged approach demonstrates that curriculum learning not only stabilizes the reinforcement learning process but also allows the model to gradually acquire both reasoning ability and refusal competence, ultimately leading to superior performance in challenging multi-modal QA tasks.

\section{Conclusion}

In this paper, we presented our solution for the KDD Cup 2025 CRAG-MM competition. For task 1, we proposed an innovative curriculum learning-based reinforcement learning approach, enabling the model to progressively and stably acquire relevant knowledge at different training stages. As a result, our system achieved a significantly lower missing rate compared to other competitors on the leaderboard (0.711 vs. above 0.8 for all others). In the manual evaluation phase, our model’s truthfulness score outperformed the second-place team by 52.38\%, demonstrating the substantial advantage of our training methodology even when using the same base model. For Task 2 and Task 3, we designed a RAG module, which leverages LLM-based query generation, retrieval, coarse ranking, and re-ranking to identify the most relevant external information. Building on the foundation established in Task 1, this module further improved our model’s performance on more challenging multi-source and multi-turn question answering tasks.

\bibliographystyle{ACM-Reference-Format}
\bibliography{sample-base}

@inproceedings{devlin2019bert,
  title={Bert: Pre-training of deep bidirectional transformers for language understanding},
  author={Devlin, Jacob and Chang, Ming-Wei and Lee, Kenton and Toutanova, Kristina},
  booktitle={Proceedings of the 2019 conference of the North American chapter of the association for computational linguistics: human language technologies, volume 1 (long and short papers)},
  pages={4171--4186},
  year={2019}
}

@article{achiam2023gpt,
  title={Gpt-4 technical report},
  author={Achiam, Josh and Adler, Steven and Agarwal, Sandhini and Ahmad, Lama and Akkaya, Ilge and Aleman, Florencia Leoni and Almeida, Diogo and Altenschmidt, Janko and Altman, Sam and Anadkat, Shyamal and others},
  journal={arXiv preprint arXiv:2303.08774},
  year={2023}
}

@article{hu2022lora,
  title={Lora: Low-rank adaptation of large language models.},
  author={Hu, Edward J and Shen, Yelong and Wallis, Phillip and Allen-Zhu, Zeyuan and Li, Yuanzhi and Wang, Shean and Wang, Lu and Chen, Weizhu and others},
  journal={ICLR},
  volume={1},
  number={2},
  pages={3},
  year={2022}
}

@article{liu2025visual,
  title={Visual-rft: Visual reinforcement fine-tuning},
  author={Liu, Ziyu and Sun, Zeyi and Zang, Yuhang and Dong, Xiaoyi and Cao, Yuhang and Duan, Haodong and Lin, Dahua and Wang, Jiaqi},
  journal={arXiv preprint arXiv:2503.01785},
  year={2025}
}

@article{zheng2025deepresearcher,
  title={Deepresearcher: Scaling deep research via reinforcement learning in real-world environments},
  author={Zheng, Yuxiang and Fu, Dayuan and Hu, Xiangkun and Cai, Xiaojie and Ye, Lyumanshan and Lu, Pengrui and Liu, Pengfei},
  journal={arXiv preprint arXiv:2504.03160},
  year={2025}
}

@inproceedings{bengio2009curriculum,
  title={Curriculum learning},
  author={Bengio, Yoshua and Louradour, J{\'e}r{\^o}me and Collobert, Ronan and Weston, Jason},
  booktitle={Proceedings of the 26th annual international conference on machine learning},
  pages={41--48},
  year={2009}
}

@inproceedings{antol2015vqa,
  title={VQA: Visual Question Answering},
  author={Antol, Stanislaw and Agrawal, Aishwarya and Lu, Jiasen and Mitchell, Margaret and Batra, Dhruv and Zitnick, C. Lawrence and Parikh, Devi},
  booktitle={Proceedings of the IEEE International Conference on Computer Vision},
  year={2015}
}

@inproceedings{ma2016,
  title={Learning to Answer Questions from Image Using Convolutional Neural Network},
  author={Ma, Lin and Lu, Zhengdong and Li, Hang},
  booktitle={AAAI Conference on Artificial Intelligence},
  year={2016}
}

@inproceedings{malinowski2015,
  title={Ask Your Neurons: A Neural-Based Approach to Answering Questions About Images},
  author={Malinowski, Mateusz and Rohrbach, Marcus and Fritz, Mario},
  booktitle={Proceedings of the IEEE International Conference on Computer Vision},
  year={2015}
}

@inproceedings{he2016,
  title={Deep Residual Learning for Image Recognition},
  author={He, Kaiming and Zhang, Xiangyu and Ren, Shaoqing and Sun, Jian},
  booktitle={Proceedings of the IEEE Conference on Computer Vision and Pattern Recognition},
  year={2016}
}

@inproceedings{anderson2018bottom,
  title={Bottom-Up and Top-Down Attention for Image Captioning and Visual Question Answering},
  author={Anderson, Peter and He, Xiaodong and Buehler, Chris and Teney, Damien and Johnson, Mark and Gould, Stephen and Zhang, Lei},
  booktitle={Proceedings of the IEEE Conference on Computer Vision and Pattern Recognition},
  year={2018}
}

@article{dosovitskiy2020,
  title={An Image is Worth 16x16 Words: Transformers for Image Recognition at Scale},
  author={Dosovitskiy, Alexey and Beyer, Lucas and Kolesnikov, Alexander and Weissenborn, Dirk and Zhai, Xiaohua and Unterthiner, Thomas and Dehghani, Mostafa and Minderer, Matthias and Heigold, Georg and Gelly, Sylvain and others},
  journal={arXiv preprint arXiv:2010.11929},
  year={2020}
}

@inproceedings{lu2019vilbert,
  title={ViLBERT: Pretraining Task-Agnostic Visiolinguistic Representations for Vision-and-Language Tasks},
  author={Lu, Jiasen and Batra, Dhruv and Parikh, Devi and Lee, Stefan},
  booktitle={Advances in Neural Information Processing Systems},
  year={2019}
}

@article{su2019,
  title={VL-BERT: Pre-training of Generic Visual-Linguistic Representations},
  author={Su, Weijie and Zhu, Xizhou and Cao, Yue and Li, Bin and Lu, Lewei and Wei, Furu and Dai, Jifeng},
  journal={arXiv preprint arXiv:1908.08530},
  year={2019}
}

@article{tan2019,
  title={LXMERT: Learning Cross-Modality Encoder Representations from Transformers},
  author={Tan, Hao and Bansal, Mohit},
  journal={arXiv preprint arXiv:1908.07490},
  year={2019}
}

@inproceedings{radford2021learning,
  title={Learning Transferable Visual Models From Natural Language Supervision},
  author={Radford, Alec and Kim, Jong Wook and Hallacy, Chris and Ramesh, Aditya and Goh, Gabriel and Agarwal, Sandhini and Sastry, Girish and Askell, Amanda and Mishkin, Pamela and Clark, Jack and others},
  booktitle={Proceedings of the International Conference on Machine Learning},
  year={2021}
}

@article{meta2023,
  title={Llama 2: Open Foundation and Fine-Tuned Chat Models},
  author={{Meta}},
  year={2023}
}

@article{bai2023,
  title={Qwen Technical Report},
  author={Bai, Jinze and others},
  year={2023}
}

@article{liu2023a,
  title={Visual Instruction Tuning},
  author={Liu, Haotian and others},
  journal={arXiv preprint},
  year={2023}
}

@article{zhu2023,
  title={MiniGPT-4: Enhancing Vision-Language Understanding with Advanced Large Language Models},
  author={Zhu, Deyao and Chen, Jun and Shen, Xiaoqian and Li, Xiang and Elhoseiny, Mohamed},
  journal={arXiv preprint arXiv:2304.10592},
  year={2023}
}

@article{wang2025crag,
  title={CRAG-MM: Multi-modal Multi-turn Comprehensive RAG Benchmark},
  author={Wang, Jiaqi and Yang, Xiao and Sun, Kai and Suresh, Parth and Sharma, Sanat and Czyzewski, Adam and Andersen, Derek and Appini, Surya and Banerjee, Arkav and Choudhary, Sajal and others},
  journal={arXiv preprint arXiv:2510.26160},
  year={2025}
}

@inproceedings{xu2023crag,
  title={CRAG: A Benchmark for Retrieval-Augmented Generation in Multi-Modal and Multi-Turn Settings},
  author={Xu, Yujia and others},
  booktitle={Proceedings of the 2023 Conference on Empirical Methods in Natural Language Processing},
  year={2023}
}

\end{document}